\documentclass{article}

     \usepackage[nonatbib,preprint]{neurips_2020}

\usepackage[utf8]{inputenc} 
\usepackage[T1]{fontenc}    
\usepackage{hyperref}       
\usepackage{url}            
\usepackage{booktabs}       
\usepackage{amsfonts}       
\usepackage{nicefrac}       
\usepackage{microtype}      
\usepackage{amssymb}
\usepackage{amsmath}
\usepackage{graphicx}
\usepackage[font=small,labelfont=bf,tableposition=top]{caption}

\title{Image-to-image Mapping with Many Domains by Sparse Attribute Transfer}

\author{%
  Matthew Amodio \\
  Yale University \\
  \And
  Rim Assouel \\
  Université de Montréal, Mila \\
  \And
  Victor Schmidt \\
  Université de Montréal, Mila \\
  \And
  Tristan Sylvain \\
  Université de Montréal, Mila \\
  \And
  Smita Krishnaswamy\thanks{Equal contribution.} \\
  Yale University \\
  \And
  Yoshua Bengio\footnotemark[1] \\
  Université de Montréal, Mila \\
}

\begin{document}

\maketitle

\begin{abstract}
  Unsupervised image-to-image translation consists of learning a pair of mappings between two domains without known pairwise correspondences between points. The current convention is to approach this task with cycle-consistent GANs: using a discriminator to encourage the generator to change the image to match the target domain, while training the generator to be inverted with another mapping. While ending up with paired inverse functions may be a good end result, enforcing this restriction at all times during training can be a hindrance to effective modeling. We propose an alternate approach that directly restricts the generator to performing a simple sparse transformation in a latent layer, motivated by recent work from cognitive neuroscience suggesting an architectural prior on representations corresponding to consciousness. Our biologically motivated approach leads to representations more amenable to transformation by disentangling high-level abstract concepts in the latent space. We demonstrate that image-to-image domain translation with many different domains can be learned more effectively with our architecturally constrained, simple transformation than with previous unconstrained architectures that rely on a cycle-consistency loss.
\end{abstract}

\section{Introduction}
In the task of unsupervised image-to-image translation, we seek a mapping between two distributions of images where we do not have a known pairwise correspondence between points. Implicit in the task is the assumption that the variation in a system with two distributions can be broken down into: 1. variation separating the source and target domains and 2. variation intrinsic to just the source or just the target domain. The goal in image-to-image translation is to remove variation along the first axis while preserving all of the other variation along the axes in the second category.

Since their advent, the field has been dominated by generative adversarial networks (GANs) based on the cycle-consistency assumption~\cite{zhu2017unpaired,almahairi2018augmented,choi2018stargan,lee2018diverse,yi2017dualgan}. With the cycle-consistency assumption, two networks are trained simultaneously (going in both directions between the two domains) and they are enforced to be inverses of each other. With a cycle-consistent GAN, the removal of the variation separating the source and target domains (from 1. above) is accomplished by the discriminator distinguishing between the target domain images and the generated images. To avoid changing other sources of variation (from 2. above), the cycle-consistency loss is employed. While invertibility may be a desirable property in some mappings, it prohibits other mappings from being learned (e.g. many-to-one relationships cannot be inverted losslessly). Moreover, empirical evidence shows that training under cycle-consistency produces unwanted restrictions, like generating a set of images with highly correlated pairwise distances to the original set of images, as measured in pixel space~\cite{benaim2017one,amodio2019travelgan}. These limitations shed light on the seemingly arbitrary difference between domains where cycle-consistent models succeed impressively versus those where they utterly fail~\cite{amodio2019travelgan}.

\begin{figure}
    \centering
    \includegraphics[width=.75\linewidth]{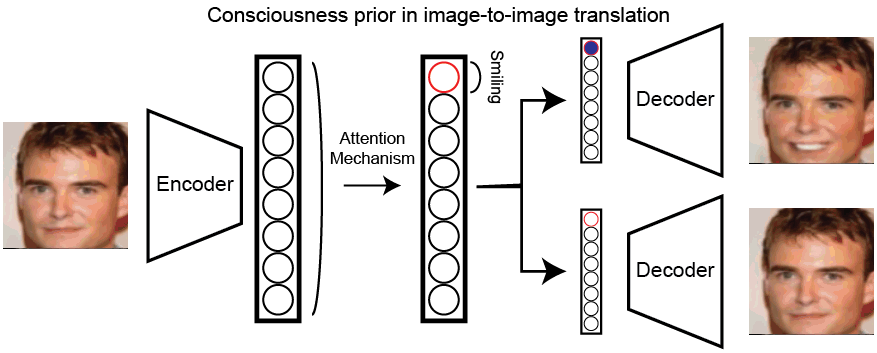}
    \caption{A conceptual figure of the consciousness prior in the context of image-to-image translation.}
    \label{fig:conceptual_fig}
\end{figure}

The problems of cycle-consistent training are especially exacerbated when mapping between many domains, for example using one model to map between domains represented by all attributes in the canonical CelebA dataset~\cite{liu2015faceattributes}. In the context of many domains, one generator network performs all of the mappings, distinguished by conditional labels. With one generator network having to be its own inverse in pixel space between many different domains, with just conditional labels to distinguish them, it comes as no surprise that existing cycle-consistency models struggle.

We offer an alternative approach: rather than a completely unrestricted architecture coupled with a cycle-consistency constraint, we directly restrict the generator architecture.

Our approach, which is called {\em semantic attribute transfer}, is built on the previously proposed notion of the consciousness prior~\cite{bengio2017consciousness}. Based off of ideas from cognitive neuroscience, the consciousness prior hypothesizes that high-level abstract concepts are stored sparsely in deep representations~\cite{baars1993cognitive,van2009stanford,dehaene2017consciousness}. Rather than performing transformations directly in pixel space, this prior supposes transformations between high-level traits can be done in a latent layer with relatively simple functions performed on sparse coordinates chosen with an attention mechanism.

Semantic attribute transfer bridges the gap between the consciousness prior, previously detailed in the context of natural language processing, and image-to-image translation. In the consciousness prior work,  consciousness is defined as a process involving the restriction of the attention of a network to changing one feature at a time. While they view this as consciousness \textit{over time}, we modify this as consciousness \textit{in abstract representation feature space}.

We restrict the translation to be a consciousness process in the following way: we eschew the cycle-consistency loss term, but instead of using an entirely unconstrained generator architecture, we only allow it to perform a single-feature transformation. We train the generator to autoencode, \textit{i.e.} when given an image as input, it produces the same image as output. Jointly, we allow an attention mechanism to select a \textit{\textbf{single}} neuron in the middle layer of the generator, perform a simple parametric change to its value, and finish the feedforward to produce a generated image (Figure~\ref{fig:conceptual_fig}).

This construction places a prior on the representation of high-level concepts in the latent layer of the generator, corresponding to each domain being stored as independent factors at this level. As proposed in the previous work on the consciousness prior, this allows each transformation to constitute a single ``conscious thought''. With this, an analogy can be made to how a human would focus consciousness on one main difference between the two distributions, and ignore all other minor differences.

Such a strictly constrained transformation function ensures only the dominant axis of variation between domains is changed, as it can only change the amount of information that can be stored in a single neuron (while also being able to autoencode from the representation). Furthermore, this obviates the need for heuristic losses frequently used in image-to-image translation like the cycle-consistency loss or the loss that enforces a network not to change input that is already in the target domain~\cite{zhu2017unpaired}. 

The main experimental findings are as follows:
\begin{enumerate}
    \item Our semantic attribute transfer model improves performance on image-to-image translation tasks with many different domains trained simultaneously, using traits from CelebA and CUB~\cite{liu2015faceattributes,WelinderEtal2010}.
    \item  The strength of our representation can be \textit{improved} by adding more domains to the mapping problem due to increasingly disentangled representations. This makes our model more scalable to increasing numbers of domains than previous work.
    \item Our results indicate that requiring cycle-consistency in training is an overly restrictive heuristic: without it our model learns inverse functions when domains are inherently one-to-one and not when otherwise.
\end{enumerate}

\section{Related Work}
\paragraph{Consciousness Prior}
The consciousness prior hypothesizes a biologically motivated representation constraint on the manipulation of high-level, abstract characteristics in the data~\cite{bengio2017consciousness}. In previous work, it was explained in terms of consciousness over time in a recurrent natural language processing model. This framed the memory of a model at time $t$ in terms of its memory at time $t-1$, using a \textit{consciousness process} $C$:
\begin{gather}
    c_{t} = C(c_{t-1}, a_t, f_t) \nonumber
\end{gather}
By restricting $a_t$ to be sparse and $f_t$ to be simple, we can simulate the model focusing on only one feature at each step.

\paragraph{StarGAN}
The StarGAN extended image-to-image translation work by considering many domains~\cite{choi2018stargan}. To avoid the combinatorial growth of having to train distinct models for every pair of domains, they use one generator network and one discriminator network that are conditioned upon the target domain. Then, using the original domain label, a cycle-consistency loss is imposed upon both directions. Otherwise, their architectures are the completely unconstrained ones commonly used in other image-to-image translation models.

\section{Model}

\begin{figure}
    \centering
    \includegraphics[width=.75\linewidth]{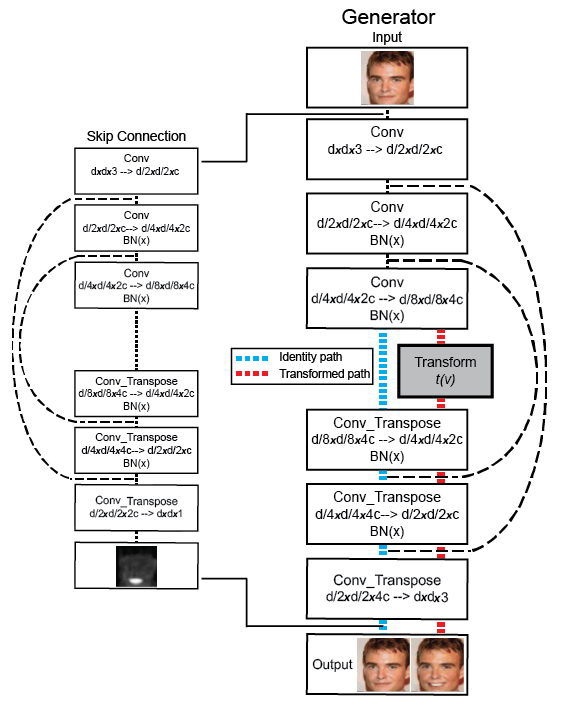}
    \caption{The generator network architecture, simultaneously trained to produce the identity (blue path) and a transformed image (red path), with only a single neuron change in the middle layer different between the two.}
    \label{fig:arch}
\end{figure}

Here we consider the multi-domain image-to-image translation problem. Let $\{x_{ij}\}_{j=1}^{n} \in X_i$ be a finite sample from domain $i=1...n_{domain}$. We seek mappings $G_i : X_{i}^c \rightarrow X_{i}$ that take points outside of domain $i$ (in the complement) and map them into domain $i$. This terminology emphasizes that we want a separate mapping for each domain $i$ that maps \textit{into} domain $i$ from anywhere else in the space. We note that the domains may not form a partition as they may overlap. We employ the generative adversarial network formulation that trains these mappings by pitting a generator against a discriminator in the standard alternating optimization process:
\begin{gather}
    \underset{G}{min} \: \underset{D}{max} \: L_{GAN} = \mathbb{E}_{x_i \sim p_{X_i}}[log D(x_i | i)] + \mathbb{E}_{x_i^c \sim p_{X_i^c}}[log(1 - D(G(x_i^c | i) | i)] \nonumber 
\end{gather}
Each mapping $G_i$, which maps \textit{into} domain $i$, shares the same network weights. The target domain to map into is selected via a conditioning label $i$. We choose this rather than using separate networks for each generator $G_i$ and each discriminator $D_i$ to facilitate scaling to large numbers of domains.

In our image-to-image domain translation context, we modify the previous definition of a \textit{consciousness process} $C$:
\begin{gather}
    c_i = C(c_j, a_i, f_i) \nonumber
\end{gather}
Instead of operating on a time axis, it is now a function that operates on a low-level representation of an image in domain $j$, and parameterized by a sparse attention mechanism $a_i$ and a simple parameterized edit $f_i$, produces a low-level representation of an image in domain $i$.

To create a generator that uses such a consciousness process, we create a novel architecture for the generator function $G(x | i)$. The conditioning label is provided to the generator by concatenating a one-hot vector to the input channel-wise. We decompose our generator into a downsampling encoder $E$ and an upsampling decoder $H$. We run our generator in two distinct modes (optimized jointly), and in the first it is simply the composition of the encoder and the decoder. In this part of the total loss we train it as an autoencoder with the usual reconstruction loss, which will result in an approximate identity function.

\begin{align}
    L_{recon} &= ||x - G(x | i)||^2 \nonumber 
\end{align}

Simultaneously, we run the generator in another form designed to perform the transformation. To do so, we place a transform $t(v)$ in between the encoder and decoder:
\begin{gather}
    G_t(x | i) = H(t(E(x))) \nonumber
\end{gather}
Crucially, by enforcing $G$ to be the identity and restricting $t$ properly, we ensure that the generating function $G_t$ conforms to the notion of the consciousness prior, by only allowing the ``consciousness'' of the generator to focus on a single element, selected via a sparse attention mechanism, and change it with a simple edit.

The transform $t(v)$ consists of two aspects: an attention mechanism which selects a feature dimension to edit, and a simple quadratic edit. We chose the simplest transformation that had sufficient expressiveness to produce domain adaptation results (an affine transformation was insufficiently expressive). Let $v \in \mathbb{R}^{d}$ be a feature vector representation in the internal layer of the generator. A domain-specific attention vector $\tilde{a_i}$ selects one of the dimensions to edit, and domain-specific parameters $s_{1i}, s_{2i}, b_i$ are used in the simple quadratic edit:
\begin{gather}
    a_i = softmax(\tilde{a_i}) \nonumber \\
    t(v) = a_i \cdot (s_{2i} \cdot v^2 + s_{1i} \cdot v + b) + (1 - a_i) \cdot v \nonumber
\end{gather}
where $\cdot$ denotes element-wise multiplication. To ensure that the attention mechanism is sparse, we add an entropy penalty to the loss:
\begin{gather}
    L_{entropy} = - \underset{ij}{\Sigma} \left ( a_{ij} \cdot log(a_{ij}) + (1 - a_{ij}) \cdot log(1 - a_{ij}) \right ) \nonumber
\end{gather}

When adapting the consciousness prior concept from natural language to images, we are faced with the additional complication of points whose feature depth is distributed across spatial locations. Thus, here we observe that while the transform $t(x)$ performs the domain translation in a sparse way \textit{with respect to the feature depth}, it does not have spatial awareness. To address this, we introduce a skip connection network to act as a mask and select the spatial coordinates for the transform to apply to. This skip connection network has a one-channel sigmoid output mask $M$ and is used in the final generator model in the following way:
\begin{gather}
    G(x | i) = M \cdot G(x | i) + (1 - M) \cdot x \nonumber
\end{gather}

The total objectives for the generator and the discriminator, respectively, are thus:
\begin{align}
    L_G &= L_{GAN} + \lambda_{recon} \cdot L_{recon} + \lambda_{entropy} \cdot L_{entropy} \nonumber \\
    L_D &= L_{GAN} \nonumber
\end{align}
with coefficients $\lambda_{recon}, \lambda_{entropy}$ controlling how much each term weighs relatively in the total.



\section{Experiments}
There are many challenges to building models in the multi-domain image-to-image context. Most significantly, many different mappings must be learned at once. In a typical image-to-image task, only two domains are considered, and there is one generator entirely dedicated to going in each direction. That means each generator's weights only need to be directed towards one set of pixel combinations (\textit{e.g.} learning weights that can only produce red coloring, thus never erroneously producing yellow coloring, because the network just tunnels all output towards the red part of the space). In this context, though, many different sets of pixel combinations all need to be generated at different times. Beyond just the challenges in generation, the adversarial discriminator also faces a more difficult task in having to distinguish between many different domains, as opposed to just one.

Our architecture, based on the U-Net framework, consists of three stride-two downsampling layers with Leaky ReLU activations and three upsampling layers with regular ReLU activations, and skip connections between downsampling layers and upsampling layers of the same dimensionality~\cite{ronneberger2015u}. We use the Adam optimizer to train with a learning rate of $0.0002$ and momentum parameters $\beta_1 = 0.5, \beta_2 = 0.9$~\cite{kingma2014adam}. Resolution $128 \times 128$ is used for inputting the images. The loss function is the standard sigmoid-based GAN loss. The discriminator's architecture is identical to the downsampling half of the generator except with a flattening, conditional projection, and classification layer added on top of the last layer.

We use several baseline models as comparisons. The most relevant method we compare to for multi-domain image-to-image translation is the StarGAN~\cite{choi2018stargan}. The StarGAN relies on a cycle-consistency loss, but is architecturally equivalent to ours, allowing us to test this form of loss against our approach which follows the consciousness prior. We compare to it with varying choices for their hyperparameters, the coefficient of the cycle-consistency and the coefficient of the identity loss, with StarGAN$^1$ being 10 and 5, StarGAN$^2$ being 10 and 0, and StarGAN$^3$ being 1 and 0, respectively. We then additionally compare to two other multi-modal cycle-consistency models, namely the MUNIT and NoiseAugmented models~\cite{huang2018multimodal,almahairi2018augmented}. We note that these models are designed for producing multiple outputs within one target domain, but do not natively handle multiple domains like we have in our context. We thus use the same approach as in the StarGAN of giving the model the target label to learn a conditional output distribution.

\subsection{CUB}
In our first experiment, we perform image-to-image translation using primary color of birds in the CUB dataset~\cite{WelinderEtal2010}. We use each primary color with more than 100 image (eight total domains).

\begin{figure}
    \centering
    \includegraphics[width=\linewidth]{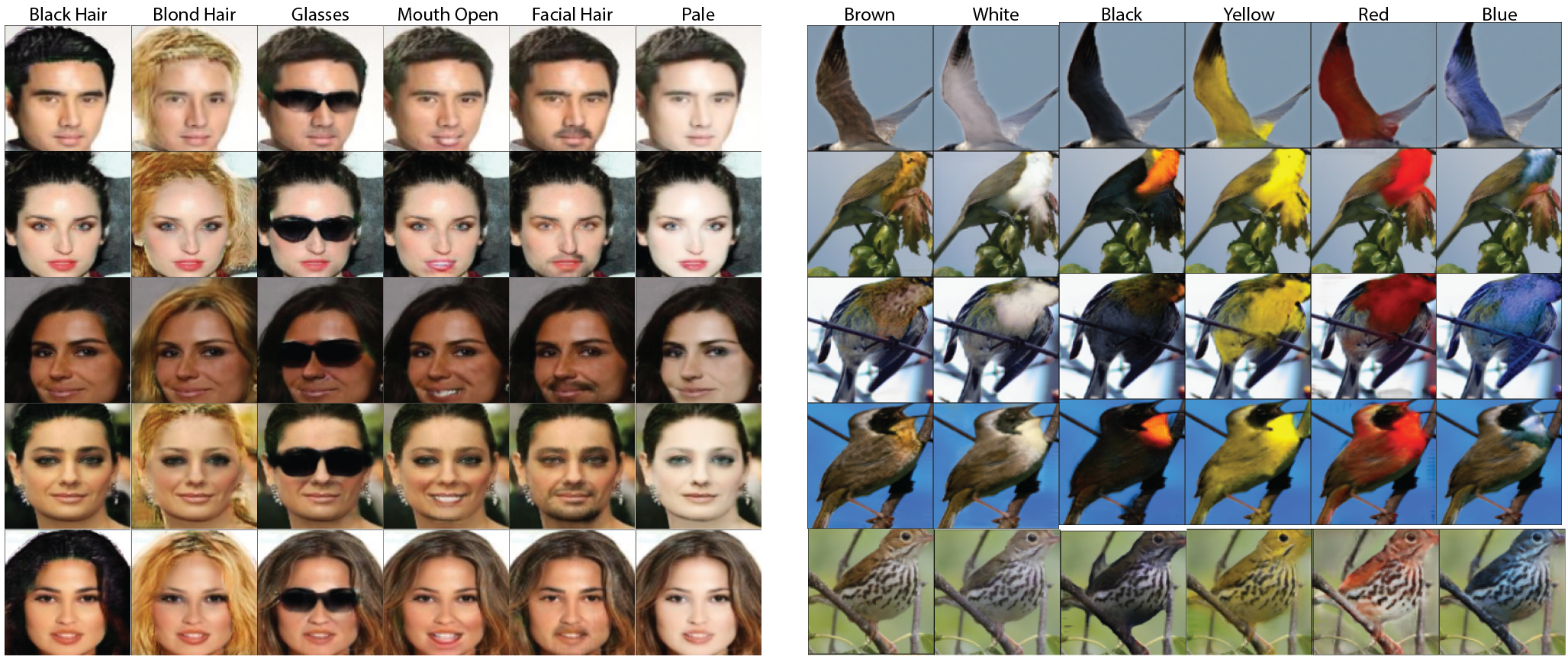}
    \caption{Samples from our model on the CelebA and CUB datasets. Individual examples are held constant as the target distribution used for the conditioning label is changed.}
    \label{fig:samples}
\end{figure}

Samples from our model with changing domains are shown in Figure~\ref{fig:samples}. We emphasize that each image is produced by changing just \textit{one neuron} in the middle layer away from the identity function. Through this, we see that the concept of primary color has been stored sparsely and simply, conforming to the hypothesis of the consciousness prior. 

To quantify the performance, we utilize the traditional measure of generative distributional accuracy, the FID score. In image-to-image translation, it is insufficient to measure image quality, as a real image is provided to the generator as input. Instead, a distributional score is necessary to measure how well the input images match the images in the target domain. For this reason, FID score, which measures distributional distance to the target distribution, is the most appropriate metric. In Table~\ref{tab:cub_meanfid}, we report the average FID score across all domains from our model and the baselines. We found our model performed significantly better than all baselines. In Figure~\ref{fig:cub_fid}, we plot the FID score per domain. Our model has a consistently low FID in each domain, with only its worst domain FID being higher than any other models' best domain FID.

\subsection{CelebA}
In the previous experiment, there were only eight domains and they were characterized by color changes only. From here we move on to a harder test of more domains that are characterized by more complicated transformations in the CelebA dataset. Because the attributes in the dataset are not necessarily mutually exclusive (e.g. an image can both be ``black-hair'' and ``glasses''), we expand the number of attributes $n_a$ into $2 \cdot n_a$ domains, where each attribute has a domain corresponding to ``has the attribute'' and a domain corresponding to ``does not have the attribute''. 

\begin{figure}
\begin{minipage}{.48\textwidth}
    \centering
    \includegraphics[width=.6\linewidth]{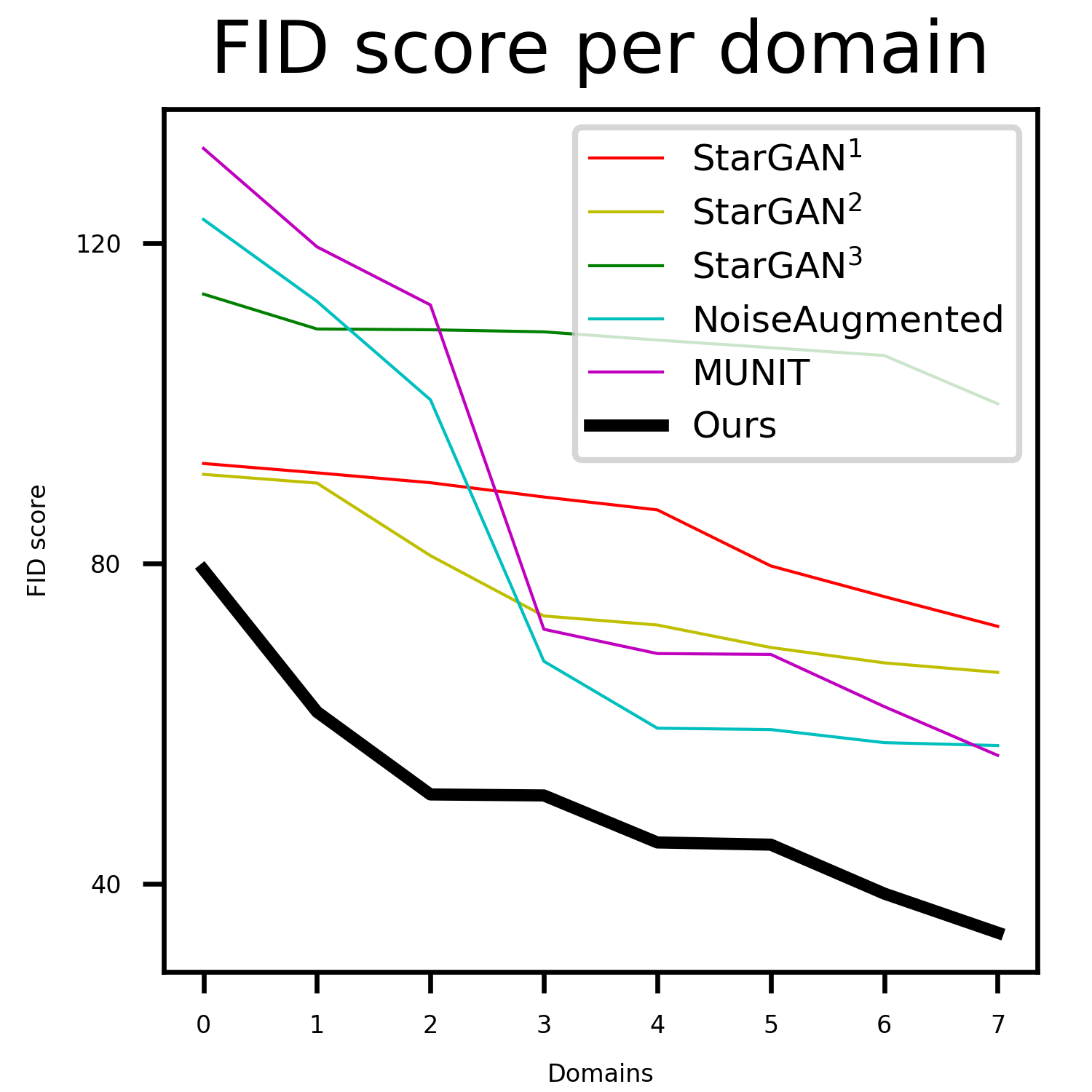}
    \caption{FID scores on the CUB dataset, per domain. The domains are ranked by FID per model.}
    \label{fig:cub_fid}
\end{minipage}\hfill%
\begin{minipage}{.48\textwidth}
    \centering
    \includegraphics[width=.6\linewidth]{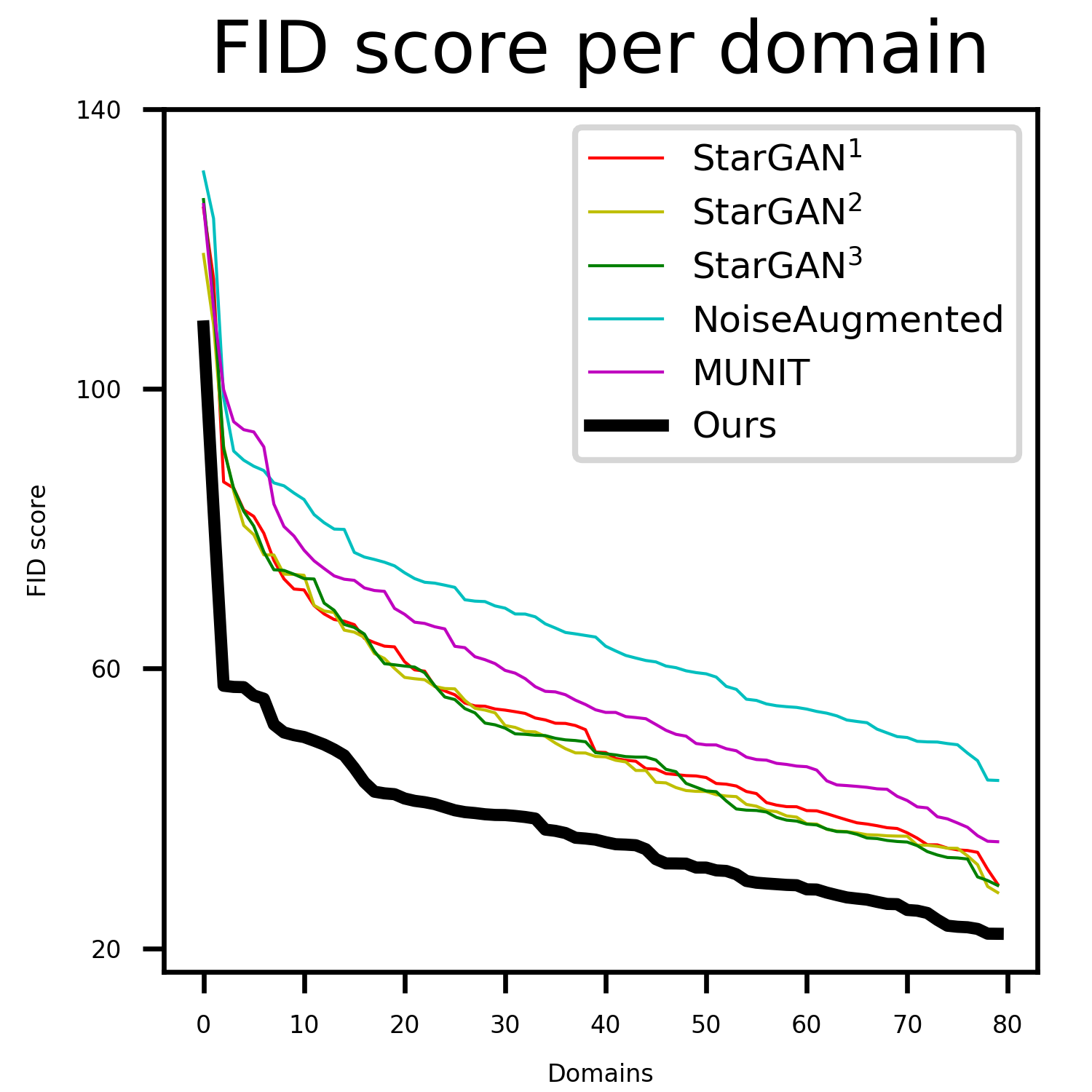}
    \caption{FID scores on the CelebA dataset, per domain. The domains are ranked by FID per model.}
    \label{fig:celeba_fid}
\end{minipage}
\end{figure}

In Figure~\ref{fig:samples}, we see samples from our model with changing target attributes, again only different from the identity function by a single simple change to \textit{one neuron} in the middle layer. For example, by being able to produce two images, one without glasses and with glasses, that only differ in their latent representation by one neuron, this provides strong empirical evidence that the abstract concept of ``glasses'' has indeed been stored sparsely and simply. As before, in Figure~\ref{fig:celeba_fid} we show the FID score per domain for each model, and in Table~\ref{tab:celeba_meanfid} we report the mean FID across all domains per model. In this case as in the previous experiment, we see our model significantly outperform the cycle-consistent models across all domains.

We next perform a series of experiments on this dataset to more thoroughly examine our model to better understand how it outperforms the best alternative StarGAN model.


\paragraph{Increasing number of domains effect}
We first evaluate the effect of increasing the number of domains that must be mapped between, by designing an experiment that isolates this effect, which we call the ``crowding effect''. As an example of the crowding effect, a model might be able to generate high quality ``glasses'' images if that is the only output domain, because it can simply learn to ignore any part of the image that is far from the eyes. However, if a model must both generate different hair colors given one condition label \textit{and} glasses given another condition label, the performance on generating glasses may suffer.

To quantify this effect, we train a model on just $Domain_i$, and measure its FID. Then, we train a new model on both $Domain_i$ and $Domain_j$, and compare the FID on $Domain_i$ from the second model to its FID in the first model. We keep doing this, adding more domains in powers of two. This gives a series of FID scores for each domain corresponding to increasing numbers of total domains in the model. We build a simple linear regression on these scores and look at the fitted slope of the regression line: an estimate of the change in FID for a given domain from adding another domain to the total mapping model.

In other words, these scores represent the change in FID for a given $Domain_i$ from making a $n$-domain mapping problem a $2n$-domain mapping problem. We report the average of these scores for each model in Table~\ref{tab:crowding}. While the cycle-consistency model has an FID that worsens as the number of domains increases, our model actually \textit{improves} its FID as the number of domains increases.

\paragraph{Latent space disentanglement}
We hypothesize one reason for our model's better generative performance is that the representation it learns and operates on better disentangles the attribute it is transforming. To investigate this, we look at the domains corresponding to ``having'' and ``not having'' each attribute, and calculate the maximal mean discrepancy (MMD) as a distributional distance between them in the deepest latent layer of each model. The distances are reported in Table~\ref{tab:disentanglement}, with the mean and standard deviation across all domains. We see that our model does indeed produce significantly more separation between points with an attribute and points without that attribute, providing evidence that our model has more effectively disentangled this variation from other underlying variation in the data.

\begin{figure}
\begin{minipage}{.48\textwidth}
    \centering
    \small
    \begin{tabular}{|c|c|}
        \hline
         & Mean FID \\
        \hline \hline
         Ours &  \textit{\textbf{50.710 +/- 13.316}} \\
         \hline
         MUNIT & 79.694 +/- 25.865 \\
         \hline
         NoiseAugmented & 86.381 +/- 27.798 \\
         \hline
         StarGAN$^1$ & 84.580 +/- 7.166 \\
         \hline
         StarGAN$^2$ & 76.433 +/- 9.181 \\
         \hline
         StarGAN$^3$ & 107.725 +/- 3.626 \\
         \hline
    \end{tabular}
    \captionof{table}{FID scores on the CUB dataset.}
    \label{tab:cub_meanfid}
\end{minipage}\hfill%
\begin{minipage}{.48\textwidth}
    \centering
    \small
    \begin{tabular}{|c|c|}
        \hline
         & Mean FID \\
        \hline \hline
         Ours &  \textit{\textbf{37.588 +/- 13.347}} \\
         \hline
         MUNIT & 58.821 +/- 18.132 \\
         \hline
         NoiseAugmented & 66.358 +/- 15.998 \\
         \hline
         StarGAN$^1$ &  52.927 +/- 17.592 \\
         \hline
         StarGAN$^2$ &  51.680 +/- 17.448 \\
         \hline
         StarGAN$^3$ &  51.937 +/- 18.914 \\
         \hline
    \end{tabular}
    \captionof{table}{FID scores on the CelebA dataset.}
    \label{tab:celeba_meanfid}
\end{minipage}
\end{figure}

\begin{table}
\begin{minipage}{.48\textwidth}
    \centering
    \small
    \begin{tabular}{|c|c|}
        \hline
         & Change in FID \\
        \hline \hline
         Our Model &  -1.622 +/- 1.023 \\
         \hline
         StarGAN & 2.404 +/- 1.212 \\
         \hline
    \end{tabular}
    \caption{Crowding effects: the change in FID for $Domain_i$ if an $n$-domain mapping problem becomes a $2n$-domain mapping problem. For our model the FID actually \textit{improves}.}
    \label{tab:crowding}
\end{minipage}\hfill%
\begin{minipage}{.48\textwidth}
    \centering
    \small
    \begin{tabular}{|c|c|}
        \hline
         & MMD \\
        \hline \hline
         Our Model &  0.440 +/- 0.074 \\
         \hline
         StarGAN & 0.118 +/- 0.081 \\
         \hline
    \end{tabular}
    \captionof{table}{MMD between the distribution of points with an attribute and the distribution without the attribute, in the deepest latent layer for each model. Our model better disentangles the attribute and separates them.}
    \label{tab:disentanglement}
\end{minipage}
\end{table}

\begin{table}
\begin{minipage}{.48\textwidth}
    \centering
    \small
    \begin{tabular}{|c|c|}
        \hline
         & Correlation \\
        \hline \hline
         Our Model &  0.847 +/- 0.163 \\
         \hline
         StarGAN & 0.976 +/- 0.009 \\
         \hline
    \end{tabular}
    \caption{Pairwise distance correlations between learned transformations on all possible domains on the CelebA dataset. Some transformations require not-near-perfect correlations.}
    \label{tab:corrs}
\end{minipage}\hfill%
\begin{minipage}{.48\textwidth}
    \centering
    \small
    \begin{tabular}{|c|c|}
        \hline
         & Mean FID \\
        \hline \hline
         Full Model &  \textit{\textbf{37.588 +/- 13.347}} \\
         \hline
         No recon loss &  39.429 +/- 14.484 \\
         \hline
         No entropy loss & 41.445 +/- 14.885 \\
         \hline
    \end{tabular}
    \captionof{table}{The ablation study results with the mean FID across all domains. The full model with all loss terms performs better than models without all of them.}
    \label{tab:ablation}
\end{minipage}
\end{table}

\paragraph{Pairwise distance correlation}
Previous work has shown that cycle-consistent mappings are constrained to learning transformations that preserve pairwise distances between points \textit{in ambient pixel space}~\cite{benaim2017one,amodio2019travelgan}. Here we perform an experiment to investigate whether this extended to the multi-domain setting. As this restriction may not always be desirable, we evaluate whether it holds in our model without a cycle-consistency loss. We take a fixed set of images, and then transform them to each target domain in turn. For each pair of domains, we calculate the correlation between the pairwise distances of points in the first domain with their representations in the other domain. Table~\ref{tab:corrs} shows the StarGAN produces high correlations in every domain (the minimum between any pair was 0.968). In our model, some domains had high correlations, while others did not (the minimum was 0.484). 

\paragraph{Cycle-consistency in our model}
While we do not enforce cycle-consistency in our model, we can evaluate whether our fully-trained model has the property. We map to and from each domain and report the average mean-squared error (MSE) between the original image and composite output after the cycle. To compare, we note that the cycle-consistent baseline had an average MSE of $0.024$, but the standard deviation was just $0.003$: this value is almost identical for every pair of domains. This makes sense, as the model is explicitly trying to minimize this term. In our model, the average MSE was $0.010$, and the standard deviation was drastically higher: $0.027$. For some domains, the reconstruction error was almost zero, while for other domains the reconstruction error was much higher than the baseline's highest. Not all desirable transformations are one-to-one and invertible: our model produces cycle-consistent mappings in some cases, and not in others.

\begin{figure}
    \centering
    \includegraphics[width=\linewidth]{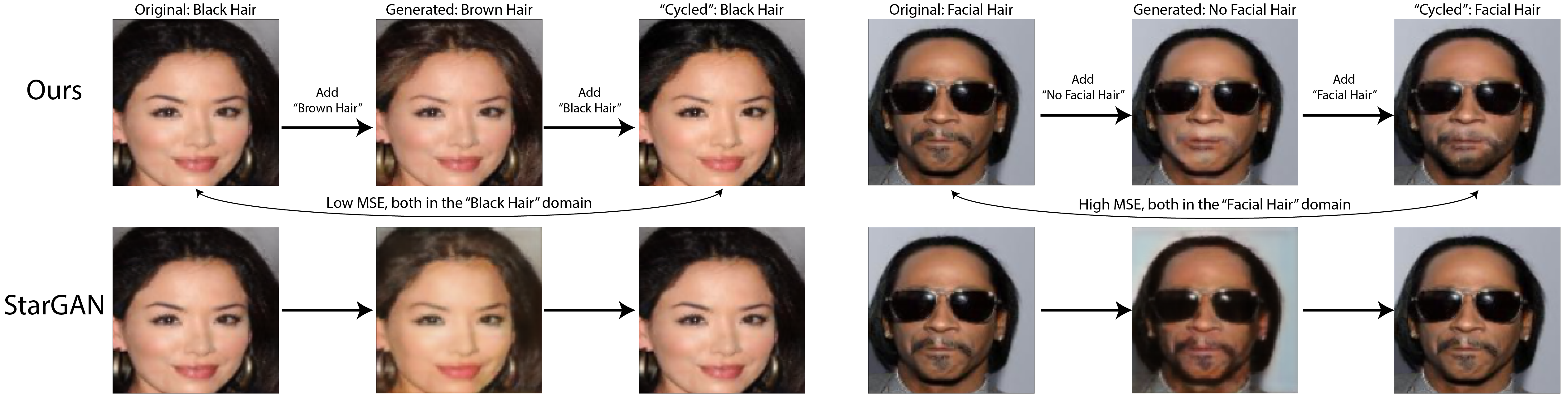}
    \caption{Some domains are inherently one-to-one and our model is cycle-consistent (despite training not with it). Other domains are many-to-one and our model is not forced to learn arbitrary one-to-one mappings. The StarGAN is prevented from changing the input at all in the latter case by the cycle-consistency restraint.}
    \label{fig:cycle}
    \vspace{-12pt}
\end{figure}

Figure~\ref{fig:cycle} shows an example of a mapping that is cycle-consistent in domains where it is natural (black hair to brown hair and back to black hair) and an example of a mapping that is \textit{not} cycle-consistent but still reasonable. An image with facial hair is mapped to the ``no facial hair'' domain and then re-mapped to the ``facial hair'' domain. The resulting image doesn't have the same facial hair as the original (it has a beard instead of the original soul patch), but is a perfectly reasonable mapping. For the StarGAN, it handles the first transformation but for the second one it is prevented from changing the input at all by the restriction of cycling back to exactly the image it started with.

\paragraph{Ablation study}
Lastly, we perform an ablation study to measure the effect of the different loss terms in our model: the reconstruction loss which forces the generator to produce the identity without the semantic attribute transform and the entropy loss which forces the attention mechanism in the transform to be sparse. In Table~\ref{tab:ablation}, we report the mean across all domains. Each loss term contributes to the ultimate performance, as the full model achieves the best score.


\section{Discussion}
Here we presented a novel framework that conforms with the notions from the recent theoretical advances in cognitive neuroscience, motivating future work in biologically motivated architectures.

\section{Broader Impact}
Our proposed method has impact on the research community, but we believe it is safe from concerns about negative ethical or societal consequences. As an architectural advance in image-to-image translation, our model does not implicate any negative consequences moreso than any other image-to-image translation model like the CycleGAN or the StarGAN. Moreover, by using pre-determined attributes in canonical datasets, we do not isolate any attribute that is not used ubiquitously in machine learning research. Many useful applications of this kind of work exist, like mapping between healthy and diseased medical imaging results.

One potential positive impact of our work is the furthering of progress towards biologically plausible and biologically motivated architectures. We believe the previous work of the consciousness prior, along with ours here, can offer the opportunity for the deep learning community to become more involved with the cognitive neuroscience community. Such a grounding in a biological science would stand to benefit the broader community in our opinion.

\bibliographystyle{abbrv}
\bibliography{references}

\end{document}